# Spelling Correction in Agglutinative Languages

Kemal Oflazer
Department of Computer Engineering and Information Science
Bilkent University
Ankara, 06533, Turkey
ko@cs.bilkent.edu.tr

**ABSTRACT**

This paper presents an approach to spelling correction in agglutinative languages that is based on two-level morphology and a dynamic programming based search algorithm. Spelling correction in agglutinative languages is significantly different than in languages like English. The concept of a word in such languages is much wider that the entries found in a dictionary, owing to productive word formation by derivational and inflectional affixations. After an overview of certain issues and relevant mathematical preliminaries, we formally present the problem and our solution. We then present results from our experiments with spelling correction in Turkish, a Ural–Altaic agglutinative language. Our results indicate that we can find the intended correct word in 95% of the cases and offer it as the first candidate in 74% of the cases, when the edit distance is 1.

## 1 Introduction

Spelling correction is an important component of any system for processing text. Creation of textual information is prone to many errors introduced by typing (human) or recognition (OCR systems) mistakes. Agglutinative languages such as Turkish or Finnish, differ from languages like English in the way lexical forms are generated. Words are formed by productive affixations of derivational and inflectional suffixes to roots or stems like, "beads-on-a-string" [14]. Furthermore, roots and suffixes (morphemes) may undergo changes at the boundaries due to various phonetic interactions. A typical nominal or a verbal root may have thousands (or even millions) of valid forms which never appear in the dictionary. For instance, we can give the following (rather exaggerated) example from Turkish:

*uygarlaştıramayabileceklerimizdenmişsinizcesine*

whose root is the adjective *uygar* (civilized).[1] The morpheme breakdown (with morphological glosses underneath) is:[2]

---

[1] This is an adverb meaning roughly "(behaving) as if you were one of those whom we might not be able to civilize."

[2] Glosses in parentheses indicate derivations not explicitly indicated by a morpheme.

| *uygar* | +*laş* | +*tır* | +*ama* | +*yabil* | +*ecek* |
|---|---|---|---|---|---|
| civilized | +AtoV | +CAUS | +NEG | +POT | +VtoA(AtoN) |
| +*ler* | +*imiz* | +*den* | +*miş* | +*siniz* | +*cesine* |
| +3PL | +POSS-1PL | +ABL(+NtoV) | +PAST | +2PL | +VtoAdv |

The portion of the word following the root consists of 11 morphemes each of which either adds further syntactic or semantic information to, or changes the part-of-speech of, the part preceding it. Though most words one uses in Turkish are considerably shorter than this, this example serves to point out the fundamental difference of the spelling checking and correction problem in such languages. Methods developed for spelling correction for languages like English (see the review by Kukich [10]) are not readily applicable to agglutinative languages.

Our prior work has mainly been on spelling checking in Turkish [12, 13], and two-level morphological analysis of Turkish [11]. In this work, we develop an algorithm for spelling correction for agglutinative languages that we have applied to Turkish. Our approach uses a two-level morphological analyzer and generator,[3] coupled with a dynamic-programming like search procedure for intelligently enumerating candidate lexical forms from a given misspelled form. In the following sections, we overview the spelling correction problem in general and in agglutinative languages, present some preliminary definitions and mathematical background and introduce an algorithm for spelling correction for agglutinative languages, and finally present results from our implementation for Turkish.

## 2  The spelling correction problem

Du and Chang [3] define the spelling correction problem as follows:

> From a set of known words (dictionary), find those words that most resemble a given (misspelled) character string.

The keyword in this definition is "resemble." It is difficult to express rigorously how two strings resemble. Generally, a distance metric is used to compare two strings. The problem then becomes that of finding those words that are neighbors of a given character string with respect to a given distance metric. There have been a number of proposals to be used as the distance metric in comparing two strings [8, 10, 15]. The most popular and widely used metrics are *q-gram* and *linear trace* based metrics. In the *q*-gram metric, two strings are compared according to the number of different substrings of length *q* they share. In the linear trace method, two strings are compared according to an *edit distance* metric which measures the extent of changes one needs to apply to one of the strings to get the other string.

---

[3]We should however emphasize that there is nothing specifically dependent in our approach to two-level morphology per se.

# 3 Spelling correction in agglutinative languages

As briefly discussed earlier, agglutinative languages have certain aspects that make the spelling correction problem substantially harder and different than that for languages like English. The expression "from a set of known words" no longer implies what is usually found in typical word list, and now means "all possible words that can be generated from a given root word by derivational and inflectional suffixes." For example, Finnish nouns have about 2000 distinct forms while Finnish verbs have about 12,000 forms ([4], pp. 59–60). The case in Turkish is also similar where nouns may have about 170 basic different forms, not counting the forms for adverbs, verbs, adjectives, or other nominal forms, generated (sometimes circularly) by derivational suffixes (Hankamer [5] gives much higher figures (in the millions) for Turkish.) If we look closely into the problem, it will not be difficult to observe that it consists of two subproblems.[4] Given a misspelled word

1. determine all the roots from the dictionary that can be the root of the misspelled word, and

2. generate (systematically) all the possible words that "resemble" the given character string, from roots identified in subproblem 1.

The first step of the problem is relatively easy because of the static structure of the root dictionary. Various techniques developed for spelling correction, say, in English can usually be applied here. We will opt not to deal with cases where a root can not be determined, especially due to total or near-total deformation.

The second step is the heart of the problem. Producing all the possible words from all the known roots requires an exhaustive generate and test search procedure.

Our approach differs from that of Aduriz et.al.[1] which also uses a morphological analysis approach. This approach is however significantly different than ours in that they mainly rely redundant two level rules to do correction while our approach is based on exploiting the morphotactics information.

## 3.1 Notation

We denote the set of the surface forms of the roots in the language[5] by $R$, and the set of lexical forms of the roots by $R_{lex}$.[6] We use $X = x_1, x_2, ..., x_m, Y = y_1, y_2, ..., y_n$ to denote strings from the alphabet of the language. $X$ will denote the surface form of the incorrect or misspelled string, and $Y$ will typically denote the surface string that is a (possibly partial) candidate word. $Y_{lex}$ will denote the lexical form of this candidate string.[7] The notation $X[i:j] = x_i, x_{i+1}, ..., x_j$ refers to the substring (from characters $i$ to $j$ inclusive) of any string $X$. If $i$ is missing, then the substring refers to the prefix of the string up to and including the $j^{th}$ character. $X[0]$ denotes empty substring and $|X|$ denotes the length of string $X$. We assume the existence of a function, $surface()$ to generate surface strings from lexical strings, i.e., $surface(Y_{lex}) = Y$. The function $surface()$ applies the constraints imposed by the automata implementing the two-level morphophonemic rules for the language.

---

[4]In this paper, we do not deal with languages that have productive prefixes.

[5]From now on, *language* will refer to an agglutinative language.

[6]Here, we are referring to the two levels of forms in the two-level morphology terminology: the *lexical form* which essentially corresponds to the structure of a word in terms of morphemes etc., and the *surface form* which is the surface realization of the lexical form as allowed by the automata implementing the two-level phonetic correspondence rules [14, 2, 9, 6].

## 3.2 Distance metrics

In both parts of the problem, we need some criteria to measure how much two strings resemble each other. Two most widely accepted and readily applicable metrics are the *q-gram distance metric* on *minimum edit distance metric*.

### 3.2.1 Q-gram distance

A $q$-gram is a substring of length $q$. The $q$-gram distance between two strings is the number of $q$-grams they do *not* have in common. For example, denoting the $q$-gram distance between two strings $X$, and $Y$, as $D_q(X,Y)$, $D_2$(ahmet,mehmet) = 3 (2-grams (bi-grams) not common to both = {ah,me,eh}), and $D_3$(ahmet,mehmet) = 3 (3-grams (tri-grams) not common to both {ahm,meh,ehm}.

### 3.2.2 Edit distance

The edit distance measures how many unit operations are necessary to convert one string into another. The unit operations are *insertion, deletion, replacement* of single character and *transposition* of two adjacent characters.

**Definition 1 (Edit Distance)** [8]

---

[7]Just to make this clear we can give an example from Turkish. For instance

    `ev+lAr+nHn` (house+PLU+GEN)

represents such a lexical form where `A` represents a low unrounded vowel (a and e in Turkish) which is unresolved for frontness, and `H` represents a high vowel (ı, i, u, and ü) which is unresolved for other features. The `+`'s indicate the morpheme boundaries. When this lexical form is processed by the generation component of a two-level morphological analyzer, the surface form obtained is:

    `evlerin`

where vowel harmony rules have resolved the `A` and the `H`, and the first `n` in the last morpheme has disappeared since the previous morpheme ends with a consonant. See Oflazer [11].

[8]This is a slight modification of edit distance formulas given by Du and Chang [3] and by Wagner and Fischer, [15].

*Given two strings $X$ and $Y$ of length $m$ and $n$ respectively, then $ed(X[m], Y[n])$[9] computed according to the recurrence below gives the minimum number of insertions, deletions, replaces and transpositions one needs to perform to convert one string to the other.*

$$
\begin{aligned}
ed(X[i+1], Y[j+1]) &= ed(X[i], Y[j]) && \text{if } x_{i+1} = y_{j+1} \\
&= 1 + min\{ed(X[i-1], Y[j-1]), && \text{if both } x_i = y_{j+1} \\
& \qquad ed(X[i+1], Y[j]), && \text{and } x_{i+1} = y_j \\
& \qquad ed(X[i], Y[j+1])\} \\
&= 1 + min\{ed(X[i], Y[j]), && \text{otherwise} \\
& \qquad ed(X[i+1], Y[j]), \\
& \qquad ed(X[i], Y[j+1])\} \\
ed(X[0], Y[j]) &= j && 1 \leq j \leq n \\
ed(X[i], Y[0]) &= i && 1 \leq i \leq m
\end{aligned}
$$

## 3.3 Recognizing and generating strings in the language

We would like to capture and abstract the behavior of a morphological generator and analyzer for the given language by two finite state automata.

**Definition 2** *A finite state generator $M_g = (P, \delta, V, S, F)$ where $P$ is a set of states, $V$ is the output alphabet (of lexical morphemes), $\delta$ is the state transition relation consisting of a set of triples $(p_i, p_j, v_k)$ indicating that the machine may traverse from state $p_i$ to state $p_j$, and output (the morpheme) $v_k$ (hence we label transition edges by $v$'s), , $S$ is the starting state, and $F$ is a set of final states, generates, all correctly formed words of the language. It should be noted that it is possible to go from one state $p_i$ to another $p_j$ by more than one transition, outputting a different morpheme. We say a string $Y_{lex}$ is generated by $M_g$, if $Y_{lex}$ is formed by concatenating, in order, the outputs of the machine as we traverse starting from $S$ to one of the states in $F$. We denote by $L(M_g)$ as the set of all lexical strings generated by $M_g$.*

$M_g$ essentially captures the morphotactics of the language, and in general may contain circular transition sequences (as is the case in Turkish). Applying the function $surface()$ to a string generated by $M_g$ will give us a valid surface string in the language. We also have a finite state recognizer $M_r$ which recognizes whether given surface strings are in the language or not. When a word in the language is input to $M_r$, if $M_r$ reaches one of its final states, the input surface word is a legal word in the language; hence $M_r$ implements a spelling checking functionality for the language. Figure 1 depicts the finite state generator defined above, where the lexical forms of the morphemes label the edges between states, and states with double circles are the final states. Typically these will be very large finite state machines with hundreds to thousands of states.

---

[9] We may occasionally drop the index of one or both arguments to indicate that we are referring to the whole string.

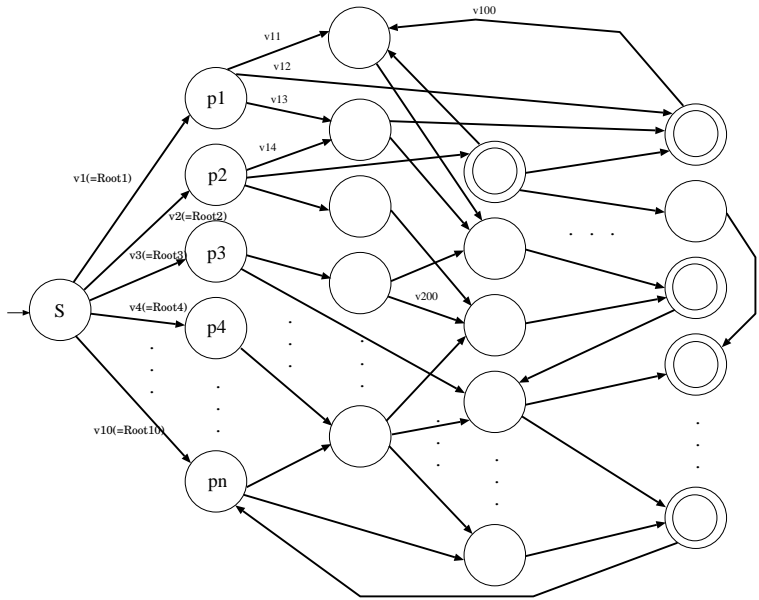

Figure 1: The finite state generator embodying the morphotactics

## 3.4 Formal description of the spelling correction problem

We can now define the spelling correction problem as:

**Definition 3** *Given an incorrect word $X$ (rejected by $M_r$), and an edit distance threshold $t$, find the solution set of possible correct words $S(X, t) = \{Y | ed(X, Y) \leq t$ and $Y = surface(Y_{lex})$ and $Y_{lex} \in L(M_g)\}$.*

In the context of the morphotactics graph shown in Figure 1, the problem can also be stated as "finding all paths from the start state (node) to all final states (nodes) such that the edit distance between the given misspelled string and the string generated by applying the *surface()* function to the concatenation of the labels of the arcs along such a path is less of equal to a given threshold." This is depicted in Figure 2. Obviously the search for such paths has to be fast.

We will now consider two subproblems of the problem.

## 3.5 Determining the root

Presenting alternatives for a given incorrect string $X$ requires determination of all possible roots. The criteria used to select roots are based on the edit distance between the (surface form) of a root and the prefixes of $X$. If any root word has an edit distance from some prefix of the misspelled word, less than the threshold $t$, then it is a candidate root. An example from Turkish makes this clear. For the misspelled Turkish word $X = kalayhlamak$, *kalayla* and *kalas* (among others) are possible roots when $t = 1$ because $ed(kalayhla, kalayla) = ed(kalay, kalas) = 1$. However, *yatay* is not a possible root

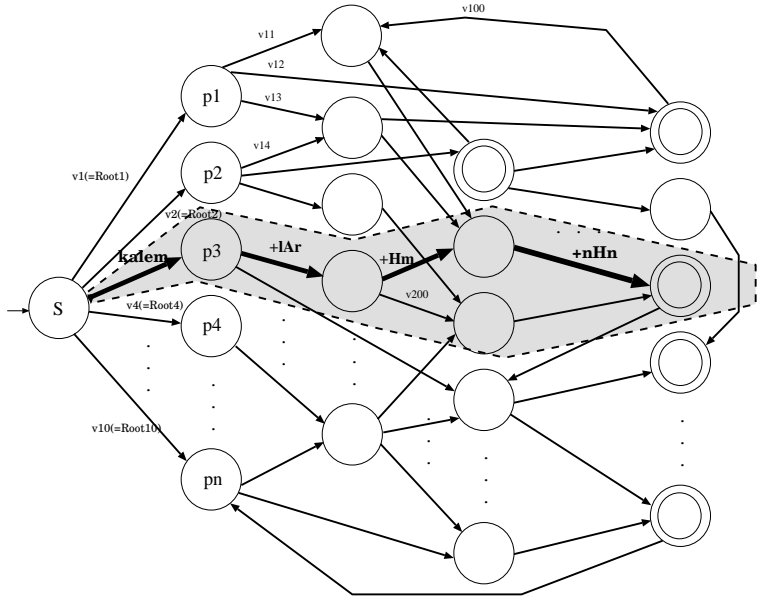

Figure 2: A path denoting a possible word in the language. The shaded area symbolizes the section of the graph to be searched.

since $ed(kala, yatay) = 3 > 1$, $ed(kalay, yatay) = 2 > 1$, and $ed(kalayh, yatay) = 3 > 1$. This observation leads to the following definition:

**Definition 4** *The set of all the possible roots for the incorrect word $X$ is, $PR(X, t) = \{r \mid ed(X[i], r) \leq t$ and $1 \leq i \leq m$ and $r \in R\}$.*

In general, the cardinality of $R$ – the set of all roots– is usually in the tens of thousands, thus one needs a fast search algorithm that works on a pre-constructed data structure for efficient determination of $PR(X, t)$. We have chosen to represent the $q$-gram information associated with root words with an inverted bit vector structure so that the bit-vector corresponding to a $q$-gram has 1's at positions corresponding to the root words containing that $q$-gram. Since the root list is static,[10] such a structure can be constructed off-line, and can be accessed randomly by using the $q$-gram as a key. Let us denote by $k$, the number of $q$-grams in a root that we would like to consult, and by $t_q$, the number of of $q$-grams we are willing to leave out and yet call the root a possible candidate root. To generate the set of such roots, we take the first $k$ $q$-grams of the incorrect word and then consider all $\binom{k}{k-t_q}$ subsets of the $(k - t_q)$ $q$-grams. For each such subset, we intersect the bit vectors corresponding to the $q$-grams in that subset. We then union the bit vectors resulting from each subset. The resulting bit vector then has 1's corresponding to root words which are "close" to a prefix of the misspelled word $X$. These roots are then filtered by the edit distance constraint in Definition 4 to compute $PR(X, t)$. The parameters $k$ and $t_q$ are in general fixed once according to the average length of the root words in the language.

---
[10] We can always deal with newly added root words in a similar fashion using different set of such bit vectors.

## 3.6 Generating candidate words from a given root

Assuming that we have a set of root words found as described above, we now have to generate words in the language having this root, that do not deviate from the given misspelled string by more than the threshold.

We will first consider solutions where the root portion of the word may be misspelled and the rest may be okay. We call such solutions as being *on the left edge of the word.*

### 3.6.1 Getting solutions on the left edge

The edit distances between an element $r$ of $PR(X, t)$ and certain prefixes of $X$ are between 0 and $t$. Sometimes, these distances are equal to $t$, which means no further mismatches between $X$ and $Y$ – the candidate string– are to be tolerated. In such cases, there is no need for further checking by generating a morpheme sequence. Just concatenating the portion of $X$ that remains after aligning $r$ with a prefix of $X$, to $r_{lex}$, and then generating the surface string will give us candidate $Y$ strings. However, determination of the alignment of the root word $r$ with $X$ is somewhat tricky because the root in $X$ may be deformed.

Let us now define a new edit distance measure between $r$, an element of $PR(X, t)$, and $X$. This is the minimum of the edit distances between $r$ and any prefix of $X$.

**Definition 5** *The prefix edit distance between $r$ and $X$ is $pred(X, r) = min\{ed(X[i], r) \mid 1 \leq i \leq m\}$.*

**Definition 6** *The set of alignment indexes of $r$ in $X$ is $index(X, r) = \{i \mid ed(X[i], r) = pred(X, r)\}$.*

For the example given before

$pred(kalayhlamak, kalayla) = 1$ and $index(kalayhlamak, kalayla) = \{8\}$
and $pred(kalayhlamak, kalas) = 1$, and $index(kalayhlamak, kalas) = \{4, 5\}$.

When $pred(X, r) = t$, the remaining part of $X$ after alignment with the root $r$ must completely occur in $Y$ after $r$ to satisfy $ed(X, Y) \leq t$. That is, $Y$ must be in the form $Y = surface(concatenate(r_{lex}, X[i + 1 : m]))$, $i \in index(X, r)$. For the example above, the candidate from root *kalayla*, is *kalaylamak*, which happens to be the correct solution and hence is accepted by $M_r$. The candidates due to *kalas* are *kalashlamak* and *kalashylamak*, both of which are rejected by $M_r$. Constructing $Y$'s for all the elements of the index set and all elements of the candidate root set, gives all possible solutions on the edge.

### 3.6.2 Generating candidate words

Getting solutions on edge will ease the computation of the correct word if the erroneous part happens to be in the root, but it does not solve the problem completely. The solution requires a generate and test probing of the graph finite-state generator $M_g$, starting with the start state $S$. We now have to find all the paths from this state to one of the final state using the roots in $PR(X,t)$, so that when the morphemes along this path are concatenated and surface string is generated, it is within an edit distance $t$ of $X$.

When the search starts morphemes are concatenated and the length of the candidate lexical string $Y_{lex}$ increases. After one step of the search, the partial surface string $Y$ is compared with a suitable prefix of $X$. In most of the cases the candidate $Y$ will deviate from these prefixes of $X$ by more than the threshold without reaching a final state, so that it can no longer lead to a viable solution. In such cases we do not consider any further transitions from that state.

The following theorem from Du and Chang [3] helps us to determine when a partial candidate $Y$ will not yield any result.

**Theorem 1** *The error matrix for all prefixes of $X$ and $Y$, is defined as $H_{m \times n}$ where $H(i,j) = ed(X[i], Y[j])$ Assume that $m \geq n$ and let $d = m - n$. Then, the sequence of elements of H, along the path $H(1,1) - H(2,1) - H(3,1) - \ldots - H(d+1,1) - H(d+2,2) - \ldots - H(m,n)$, are non-decreasing.*

**Proof:** See Du and Chang [3].

Theorem 1 determines a non-decreasing path in error distance matrix $H_{m \times n}$. This is not exactly what we need since the theorem requires that the length of the candidate string $Y$ be known. In our case, we know that this length has to be in the range $m - t$ to $m + t$ for $Y$ to be a candidate.

### 3.6.3 Limiting search during word generation

Due to the limitation above, we can not cut a branch of the search by looking at only a single path in $H_{m \times n}$ as defined in Theorem 1. First we construct $H$ for the current (possibly partial) $Y$, then consider column $n$ ($n$ being the current length of $Y$), and then find the minimum of the edit distance values along this column between rows $n - t$ and $n + t$ inclusive. If this value exceeds the threshold $t$, then there is no point in further pursuing this path, i.e., this $Y$ will not lead to any solution. Formally, we define a cut-off distance metric:

**Definition 7 (Cut-off distance)**

$$cuted(X[m], Y[n]) = \begin{cases} min\{H[i,n] \mid 1 \leq i \leq n+t\} & \text{if } n < t \\ min\{H[i,n] \mid n-t \leq i \leq n+t\} & \text{if } t < n \leq m \\ min\{H[i,n] \mid n-t \leq i \leq m\} & \text{if } m < n \leq m+t \\ n-m & \text{if } m+t \leq n \end{cases}$$

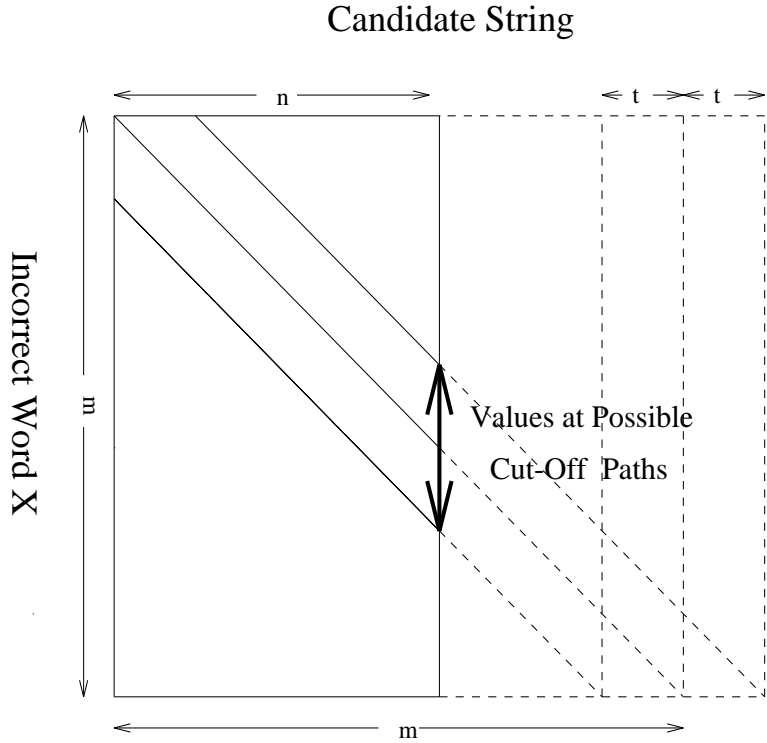

Figure 3: Determination of Cut-Off Paths in $H_{m \times n}$

The idea is similar to $pred(X,r)$ defined earlier, in that prefixes of $X$ are again considered. If the cut-off edit distance between $X$ and the current $Y$ does not exceed the threshold, further transitions along from the state in $M_g$ currently reached by $Y_{lex}$, have to be pursued.

After these observations we can state our algorithm for word generation, by searching the morphotactic graph, as follows:

```
Compute PR(X,t)
Initialize C(X,t) to the empty set
for all r ∈ PR(X,t)
/* push root and note to start search on to the stack */
  PUSH((r_lex, p_{r_lex}))
  while stack not empty
    POP((Y_lex,p_i)) /* pop the next state to check */
    for all p_j (p_i,p_j,v) ∈ δ
        Y = surface(Y_lex) /* n is the current length of Y */
        if cuted(X[m],Y[n]) ≤ t
            Push((concat(Y_lex,v),p_j))
        if ed(X[m],Y[n]) ≤ t and p_j ∈ F,
            then insert Y into C(X,t)
```

**Theorem 2** *The algorithm above produces exactly the solution set $C(X,t)$ when $PR(X,t)$ is given.*

**Proof:** Every element of $PR(X,t)$ is pushed into the stack. If $Y \in C(X,t)$ then $Y_{lex} \in L(M_g)$, that is, there is sequence of states in $M_g$ $S \xrightarrow{r_{lex}} p_{i_1} \xrightarrow{v_{i_1}} p_{i_2} \xrightarrow{v_{i_2}} p_{i_3} \cdots p_{i_{k-1}} \xrightarrow{v_{i_{k-1}}} p_{i_k}$ ($p_{i_k} \in F$) so that $Y = surface(concat(r_{lex}, v_{i_1}, v_{i_2}, \ldots, v_{i_k}))$ and for all $Y_j = surface(concat(r_{lex}, v_{i_1}, v_{i_2}, \ldots, v_{i_j}))$ $1 \leq j \leq k$ we have $cuted(X, Y_j) \leq t$.

## 3.7 Changes to the left of the morpheme boundary

In during the affixation process, some characters to the left of the morpheme boundary may be deleted or modified, though such modification will not be reflected to the partial surface form until a subsequent morpheme is added.[11] For example in Turkish, one can have a situation where the lexical form `gel+AcAk+Hm` will have the surface form `geleceğim`, yet one may not know when the second morpheme is added to the first morpheme (the root) the last `k` gets changed to a glide `ğ`, when a third morpheme is added. To handle these cases, for morphemes ending in (possibly a sequence of) characters that may undergo such changes, we can temporarily increase the threshold accordingly during edit distance matches.

# 4 Ranking the Candidate Solutions

An essential part of the spelling correction problem is the ranking of the candidate solutions. Candidate solutions can be ordered by increasing edit distance to the misspelled string. But when the number of solutions with the same edit distance is large, it is difficult to choose some subset of meaningful solutions to present the user. The problem is further complicated by the fact that the correct solution is usually determined by syntactic and semantic context and is dependent at least on the relative frequency of usage of the root words.

We have opted used a model of spelling errors based on certain statistics we have about types of spelling errors people have made in typed Turkish text. Our observation from our sample of misspelled words is that 23.1% of misspelled strings contain replacement errors, 22.2% contain a deletion, 17.3% contain an addition and 3.3% contain transposition errors. However the most dominating error type within replacements (with 34%) is the replacement of ş-s, ç-c, ı-i ö-o,ü-u, a-e pairs– all except the last one being the result of typing Turkish using a non-Turkish keyboard lacking Turkish characters or composing Turkish characters in complicated ways.

These results give us about the heuristic that we can use in ranking. First we give high priority to solutions that can be converted to misspelled string by replacement (especially as above). Then we must prefer longer solutions because deletion and replacement of characters occurs more frequently. Transpositions are of lower priority as the frequency of this error is very low in the statistics.

---

[11] We assume that the changes induced on the surface form by a new morpheme affect a very small postfix of the stem constructed so far.

## 5 Results from experiments with spelling correction in Turkish

We first present a spelling correction example from our implementation where we used bi-grams ($q = 2$), and we chose $k$ as 3 and $t_q$ as 2.

**EXAMPLE**

```
Misspelled word:        çaışmalarıyla
Threshold t:            2
Solutions on left edge: yazışmalarıyla         yatışmalarıyla
                        yapışmalarıyla         yakışmalarıyla
                        takışmalarıyla         sayışmalarıyla
                        mayışmalarıyla         katışmalarıyla
                        kapışmalarıyla         kakışmalarıyla
                        kaşışmalarıyla         çıkışmalarıyla
Candidate Roots:[12]    çağ çakı çal çalı çam çan çap çar çat çatı çav çay
                        çağ çak çakış çal çalış çap çat çatış çav

Solutions:[13]          Lexical                Surface
------------------------------------------------------------------------
Edit distance 1         çat+Hş+mA+lArH+ylA     çatışmalarıyla
                        çap+Hş+mA+lArH+ylA     çapışmalarıyla
                        çalış+mA+lArH+ylA      çalışmalarıyla (correct form)
Edit Distance 2         çav+mA+lArH+ylA        çavmalarıyla
                        çav+Hş+mA+lAr+Hm+ylA   çavışmalarımla
                        çav+Hş+mA+lAr+Hn+ylA   çavışmalarınla
                        çav+Hş+mA+lArH+ysA     çavışmalarıysa
                        çat+Hl+mA+lArH+ylA     çatılmalarıyla
                        ...                    ...
                        çat+mA+lArH+ylA        çatmalarıyla
```

The algorithm described above was tested on a set of 141 randomly selected incorrect words from Turkish text. Among these misspelled words, 14% had edit distance of 2, and the remaining 86% had edit distance 1, to their intended correct form. The morphological analyzer and generator that we used was our two-level specification for Turkish [11], developed using the PC-KIMMO system. This system has a rather comprehensive coverage of Turkish morphology and uses a root lexicon of about 24,000 words. It is, however, rather slow and can analyze only about 2 forms per second and can generate about 50 forms a second on Sun Sparcstations. So, instead of using timings, we counted the number of times the morphological analyzer and generator, and the edit distance computations, were called as these were the most expensive operations our algorithm.[14]

---

[12]The duplicate entries in the list of candidate roots for the example, are in fact not duplicate; they have different part-of-speech categories and hence different morphotactics.

[13]A small subset of the whole solution set is given here.

These statistics show the average number of morphological recognitions and generations, and the edit distance operations required, and the number of correct solutions offered *per misspelled input word*. The last column indicates the percentage of cases the intended correct form was found. The results in Table 1 are for threshold $t = 1$ and the results in Table 2 are for threshold $t = 2$. In both cases, bi-grams were used with $t_q = 2$. We varied $k$ (which determines how many bi-grams from the beginning of the incorrect word are to be considered,) between 3 and 5. This range was considered because according to some limited statistics we have on Turkish text, the average root length is about 4.5 characters. Choosing $k = 3$ allows more deformed roots to be handled at the expense of more computation, while choosing $k = 5$ sometime will not find roots with minor deformations but it runs faster.

Table 1: Average number of operations per misspelled word, for $t = 1$

| $k$ | Recognitions | Generations | Edit Distance Operations | Solutions Offered | % Accuracy |
|---|---|---|---|---|---|
| 3 | 30.9 | 311.2 | 2498.4 | 3.6 | 95.1 |
| 4 | 10.4 | 194.7 | 1068.8 | 2.4 | 78.2 |
| 5 | 3.9 | 88.5 | 471.7 | 1.5 | 54.0 |

The ranking procedure was tested on the similar set of data. Only the size of test data was increased but the percentages among the type and values of edit distances remained essentially the same. The results of the performance of the ranking procedure are given

---

[14]Although, our PC-KIMMO based morphological analyzer and generator that we have used for this study is rather slow, we have now ported our morphological analyzer system to the XEROX TWOL system by Karttunen [7], and intend to integrate it to our system. This system can recognize and generate Turkish forms in about a millisecond on Sun Sparcstations. With this system it will be possible to generate all solutions in about 1 to 2 seconds for $t = 1$ and in a few seconds for $t = 2$, on Sun Sparcstations.

Table 2: Average number of operations per misspelled word, for $t = 2$

| $k$ | Recognitions | Generations | Edit Distance Operations | Solutions Offered | % Accuracy |
|---|---|---|---|---|---|
| 3 | 108.4 | 4462.0 | 20680.4 | 52.0 | 95.1 |
| 4 | 46.5 | 2247.8 | 10386.6 | 35.5 | 78.2 |
| 5 | 13.6 | 817.1 | 3799.9 | 20.3 | 54.0 |

Table 3: The performance of the ranking procedure

| Edit Dist. | Given in First Pos. | Given | Not Given |
|---|---|---|---|
| 1 | 75.8% | 20.7% | 3.5% |
| 2 | 28.2% | 51.2% | 20.6% |
| 3-4 | 5.5% | 25.0% | 70.5% |

in the Table 3.

# 6 Conclusions

This paper has presented a spelling correction algorithm for agglutinative languages that is based on a two-level morphological generator and analyzer, and a intelligent generate and test search procedure. The algorithm uses a $q$-gram based approach to determine the candidate roots words, and then from each root word, generates valid forms in the language, that are guaranteed not to deviate from the given misspelled string by more than a threshold, using morphological generator. We have applied this approach to Turkish, and our results indicate that we can find the intended correct word in 95% of the cases and offer it as the first candidate in 74% of the cases, when the edit distance is 1. We feel that using $k = 3$ and $t = 1$, we get a satisfactory (functional) performance for Turkish. We can certainly improve on the ranking results by incorporating root usage statistics.

# 7 Acknowledgments

This work was supported in part by NATO Science for Stability Project Grant TU-LANGUAGE. Cemaleddin Güzey implemented and tested the algorithms presented here.